\pdfoutput=1
\documentclass[11pt]{article}

\usepackage{acl}

\usepackage{times}
\usepackage{latexsym}
\usepackage[T1]{fontenc}
\usepackage[utf8]{inputenc}
\usepackage{microtype}
\usepackage{inconsolata}
\usepackage{graphicx}
\usepackage{caption}
\usepackage{amsmath}
\usepackage{amssymb}
\usepackage{comment}
\usepackage{booktabs}
\usepackage{hyperref}
\usepackage[overload]{empheq}
\usepackage{makecell}
\usepackage{xcolor,colortbl}
\usepackage[ruled,vlined]{algorithm2e}
\usepackage{array,multirow}
\usepackage{comment}
\usepackage{pifont}
\usepackage{siunitx}
\usepackage{url}
\usepackage{enumitem}
\usepackage{arydshln}
\usepackage{xcolor}
\usepackage{soul}
\newcommand{\linebrcelll}[2][l]{\begin{tabular}[#1]{@{}l@{}}#2\end{tabular}}
\newcommand{\linebrcellc}[2][c]{\begin{tabular}[#1]{@{}c@{}}#2\end{tabular}}

\title{On the Robustness of Agentic Function Calling}

\author{Ella Rabinovich \\
  IBM Research \\
  \texttt{ella.rabinovich1@ibm.com} \\\And
  Ateret Anaby-Tavor \\
  IBM Research \\
  \texttt{atereta@il.ibm.com} \\}

\begin{document}
\maketitle
\begin{abstract}
Large Language Models (LLMs) are increasingly acting as autonomous agents, with function calling (FC) capabilities enabling them to invoke specific tools for tasks. While prior research has primarily focused on improving FC accuracy, little attention has been given to the robustness of these agents to perturbations in their input.
We introduce a benchmark assessing FC robustness in two key areas: resilience to naturalistic query variations, and stability in function calling when the toolkit expands with semantically related tools. Evaluating best-performing FC models on a carefully expanded subset of the Berkeley function calling leaderboard (BFCL), we identify critical weaknesses in existing evaluation methodologies, and highlight areas for improvement in real-world agentic deployments.

\end{abstract}

\section{Introduction}
\label{sec:introduction}

%\ella{we should mention two works addressing the robustness aspects: "toolsandbox: A Stateful, Conversational, Interactive Evaluation Benchmark for llm Tool Use Capabilities" and "RoTBench: A Multi-Level Benchmark for Evaluating the Robustness of Large Language Models in Tool Learning"; the first work make attempts into the right direction, but report very inconclusive results, e.g., improvement per distraction tools or parameter description removal... they also make unreasonable changes to function names and use similarity metric that encapsulates too many things, making it difficult to study the effect of modifications in isolation, it's not clear to what scenario tool augmentation was applied (out of 5-6), and on average the scores are higher then the mean; since evaluation reflects the full-path-dag-milestones-minefields similarity, it's hard to say something pertaining to tool selection; the second work perform mostly unrealistic modifications to tool names and parameters...}

%\ella{our work is the first to experiment with perturbations of user query and test their effect on function calling accuracy, yielding insights}

Large Language Models (LLMs) are reshaping artificial intelligence, shifting from static language processors to dynamic, task-oriented \textit{agents} capable of planning, executing, and refining their actions. These agents hold the potential for transformative applications across various domains, including healthcare \citep{abbasian2023conversational, mehandru2024evaluating}, finance \citep{li-etal-2024-cryptotrade, xiao2024tradingagents, ding2024large}, education \cite{yang2024content, xu2024eduagent}, and customer support \cite{huang2024crmarena, rome2024ask}. LLM agents have been revolutionarily positioned as routing systems that can act independently, make decisions and perform tasks with minimal human intervention.

\begin{figure}[h!]
\centering
\includegraphics[width=1.0\columnwidth]{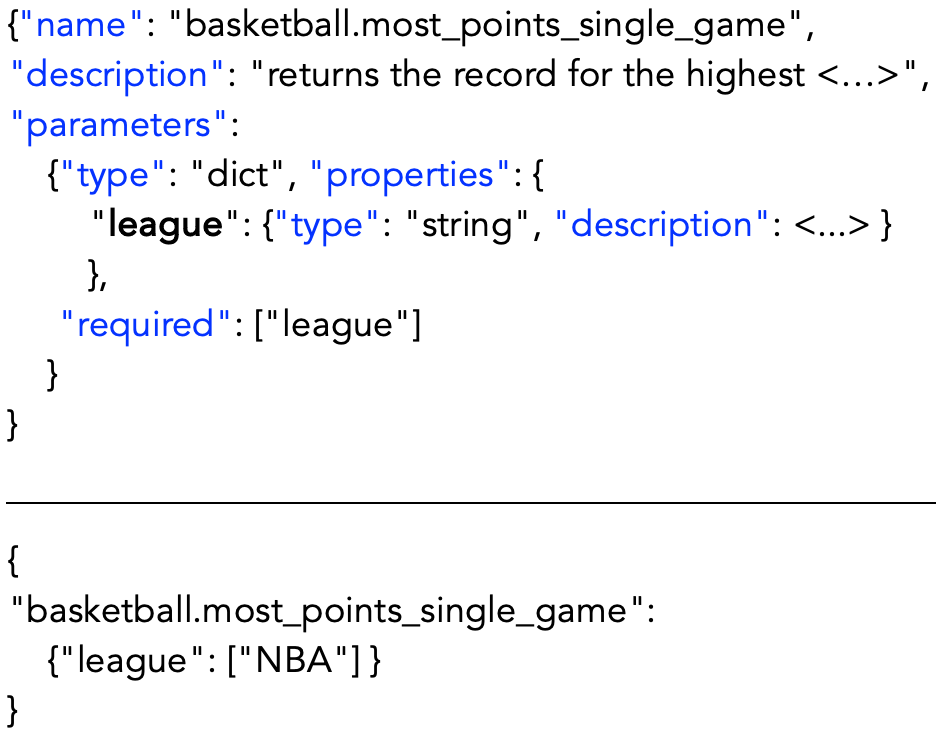}
\caption{Compact function definition example (top), and agent's output, triggering the function call with assigned parameter values (bottom), per user request "What is the record for the highest number of points scored by a single player in an NBA game?".}
%\vspace{-0.05in}
\label{fig:fc-example}
\end{figure}

\paragraph{Agentic Function Calling} Function calling (FC), the process by which an agent autonomously selects and invokes a specific function to retrieve information or execute a task, serves as a fundamental building block of an agentic system. In this context, a full execution trajectory can be seen as a complex, multi-turn (i.e., involving user interaction) sequence of function calls, ultimately achieving a given goal.
Models specifically optimized for FC are typically designed to generate a function call in response to a natural-language user request \citep{bai2023qwen, dubey2024llama, zhang2024xlam}. The function (also known as a tool) is chosen from a predefined "toolkit"---a compact set of function descriptions\footnote{Descriptions are often provided in the \texttt{json} format.}---provided as part of the model's prompt. The agent is expected to produce a syntactically correct tool invocation, ensuring that parameter values are appropriately assigned to function arguments (a process known as slot filling). For instance, given the query, "What is the record for the highest number of points scored by a single player in an NBA game?" and the compact \texttt{json} tool description in Figure~\ref{fig:fc-example} (top), the model is expected to generate the invocation code shown in Figure~\ref{fig:fc-example} (bottom). 
Several datasets and evaluation methodologies have been proposed to assess LLMs' function calling capabilities \citep{patil2023gorilla, liu2024apigen}, and various benchmarks have been created for evaluating a range of FC scenarios, \href{https://gorilla.cs.berkeley.edu/leaderboard.html}{BFCL leaderboard} \citep{patil2023gorilla} among the most prominent ones. 

\begin{comment}
\small{
\texttt{\{
"name": "basketball.most\_points\_single\_season", 
"description": "returns the record for the <...>", 
"parameters": 
    \{"type": "dict", "properties": 
        \{"league": \{"type": "string", "description": "<...>"\}\}, 
    "required": ["league"]
    \}
\}}

\texttt{\{
"basketball.most\_points\_single\_game": 
\{"league": ["NBA"]\}
\}}
}
\end{comment}

\paragraph{Robustness of Large Language Models} In the context of the more "traditional" LLM usage, a \textit{model robustness} quantifies an LLM's ability to generate semantically equivalent outputs, given semantically equivalent inputs \citep{raj2022measuring, rabinovich2023predicting, ackerman-etal-2024-novel}. Robustness benchmarks assess, among other factors, how well LLMs handle naturally-occurring, non-malicious perturbations in user input, such as paraphrased questions in a QA task, typos, variations in punctuation, whitespace, or diacritics. Extending this notion to agentic FC would require a model to produce an equivalent tool invocation despite naturalistic, yet, strictly meaning-preserving, perturbations in the input query. Considering Figure~\ref{fig:fc-example}, a semantically equivalent paraphrase "What is the highest number of points ever scored by a single player in an NBA game?" should result in the same tool invocation as the original request.

Despite its clear practical significance, research on the robustness of agentic function calling remains sparse, with only two studies, to the best of our knowledge, examining agent resilience to modifications in tool descriptions. \citet{ye2024rotbench} introduce a series of increasingly aggressive alterations to \textit{function names}, \textit{parameter names}, and their \textit{descriptions} -- to the point where a tool (or a parameter) name (or description) becomes arbitrary or entirely uninformative about its functionality. Similarly, \citet{lu2024toolsandbox} conduct multiple interventions, including tool distractions, within a different evaluation framework that evaluates tool sequencing at the \textit{system} rather than \textit{function} level. While these studies offer valuable insights, they provide limited evidence on agent resilience to real-world perturbations, as system developers typically exert \textit{substantial control} over the faithfulness and level of detail in function and parameter names, along with their descriptions. 

Moreover, a typical "toolkit" (the list of available functions) in these studies is limited to a single tool or a small number of unrelated tools. A realistic scenario may involve a system specification with thousands of available tools,\footnote{A software engineering (SWE) agent fixing git issues, has access to about 1.2K tools exposed through \href{https://github.com/github/rest-api-description/tree/main/descriptions-next/api.github.com}{github docs}.} which in practice is normally reduced to top-K most relevant function definitions through a shortlisting module \citep{qin2023toolllm}, such as semantic search over the set of tools, towards constructing the context (here, prompt) of a FC agent. In the example toolkit in Figure~\ref{fig:fc-example} (top), additional tools may include:\\ \texttt{basketball.most\_points\_career(), basketball.most\_points\_single\_season(), basketball.game\_stats()}.

\paragraph{Contribution} We focus on two aspects of robustness, capturing input variations that can be expected in real-world agentic deployments but are \textit{not easily controlled} by a developer: (1) generating meaning-preserving rephrasings of user requests and (2) expanding the toolkit to include a set of semantically related tools that are likely to be shortlisted by a selection module. Using one of the (single-turn) challenging BFCL \citep{patil2023gorilla} test sets as our starting point, we first carefully build a benchmark dataset, comprising variations pertaining to the two aforementioned aspects (Section~\ref{sec:dataset}). Next, we evaluate the robustness of several best-performing LLMs\footnote{According to the \href{https://gorilla.cs.berkeley.edu/leaderboard.html}{BFCL leaderboard} (Jan 2025).}, and discuss the breakdown of failures, highlighting (among others) prominent weaknesses of the existing agentic FC evaluation benchmarks (Section~\ref{sec:experiments}). Our benchmark data is available at \url{https://huggingface.co/datasets/ibm-research/BFCL-FC-robustness}.

\section{Dataset Generation}
\label{sec:dataset}

\begin{figure*}[h!]
\centering
\includegraphics[width=0.92\textwidth]{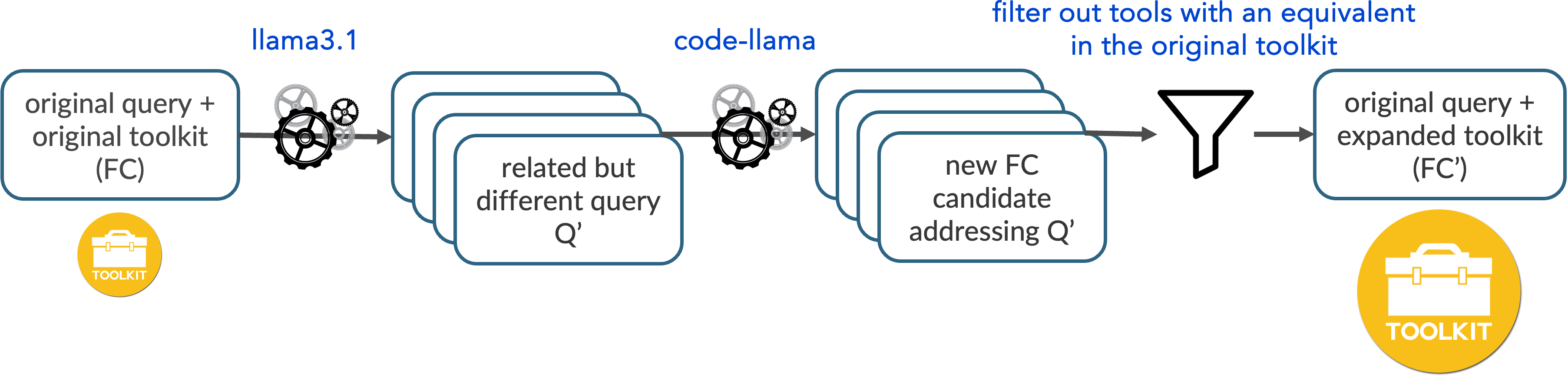}
%\vspace{-0.1in}
\caption{A toolkit expansion steps: (1) request variants are generated using the LLama3.1-70B model \citep{dubey2024llama}, (2) function \texttt{json} definitions for executing these requests are generated using the Code-Llama-13B model \citep{roziere2023code}, and a filtering step (3) is applied to filter out tools semantically identical to any of the original functions. The process is completed when the expanded toolkit is created for testing the original query.}
\label{fig:toolkit-expansion-flow}
\end{figure*}

\begin{table*}[h!]
\centering
\resizebox{\textwidth}{!}{
\begin{tabular}{l|p{14.0cm}}
original request & What is the record for the most points scored by a single player in an NBA game? \\ \hline
original toolkit & basketball.most\_points\_single\_game(...) \\ \hline \hline
request variants & \linebrcelll{Who holds the record for the highest number of assists made by a female basketball player? \\
What is the longest winning streak in NBA history? \\ 
... } \\ \hline
additional tools & \linebrcelll{
%basketball.most\_points\_single\_season(...) \\
basketball.most\_points\_career(...)\\
basketball.records\_history(...)\\
...
%basketball.game\_stats(...)
}
\end{tabular}
}
%\vspace{-0.1in}
\caption{A toolkit expansion steps: request variants and additional tools addressing those variants.}
\label{tbl:toolkit-expansion-example}
\end{table*}

\begin{comment}
\begin{table}[h!]
\centering
\resizebox{1.0\columnwidth}{!}{
\begin{tabular}{l|rrr} 
dataset & queries & M(tools) before & M(tools) after \\  \hline
multiple & ... & ... & ...\\
multiple parallel & ... & ... & ... \\
simple & ... & ... & ... \\
\end{tabular}
}
%\vspace{-0.05in}
\caption{Dataset expansion statistics.}  
\label{tbl:datasets_stats}
\end{table}
\end{comment}

We next provide details on the generation of our benchmark dataset. Specifically, we describe the creation of (1) meaning-preserving rephrasings of user requests and (2) expanding the toolkit to include a set of semantically related tools.

\subsection{User Query Perturbations}

Building on the study by \citet{ackerman-etal-2024-novel}, who tested LLMs' sensitivity to paraphrased user queries in the QA and classification settings, we investigate whether agents' FC capabilities remain robust to meaning-preserving variations in user requests. Here, the task presents additional challenge, as the rewording must strictly maintain precise parameter values to ensure accurate slot filling for the sake of evaluation. For instance, the request "Calculate the depreciated value of a property costing \$200,000 with an annual depreciation rate of 3\% for 5 years." can be safely rephrased as "Determine the value of a \$200,000 asset which loses 3 percent of its worth each year, after five years." Contemporary LLMs handle this task effectively, and we used the Llama3.1-70B model \citep{dubey2024llama}, with appropriate prompting and in-context learning. A manual review of 50 examples by one of the authors revealed no instances of semantic drift or parameter misalignment. Appendix~\ref{sec:appendix-a} provides details on the prompt used for this task.

A substantial portion of the paraphrases targeted named entities, which are natural candidates for surface form variability. For instance, the user query "What is the humidity level in \texttt{Miami,Florida} in the upcoming 7 days?" was rephrased as "How will the humidity levels change over the next seven days in \texttt{Miami,FL}?". These seemingly minor modifications led to a notable drop in benchmark performance -- we analyze and interpret this decline, and propose strategies to mitigate it in Section~\ref{sec:experiments}.

\subsection{Expanding Agent's Toolkit}

Aiming at expanding the (originally) "thin" agent's toolkit, simulating the scenario where function definitions are retrieved by a shortlister, we follow the steps illustrated in Figure~\ref{fig:toolkit-expansion-flow} and outlined in Table~\ref{tbl:toolkit-expansion-example}.

%\begin{enumerate}[label=(\arabic*)] %[leftmargin=*]
%\end{enumerate}
\vspace{-0.1in}
\paragraph{\textnormal{(1)}} We generate \textit{related yet different} request variants using the Llama3.1-70B model \citep{dubey2024llama}, see Appendix~\ref{sec:appendix-b} for the detailed prompt.
\vspace{-0.1in}
\paragraph{\textnormal{(2)}} For each request variant, a tool definition is generated to enable request fulfillment. Here, we used the CodeLlama-13B model \citep{roziere2023code} with a carefully designed prompt and few-shot examples, ensuring that the generated definitions conform not only to the required \texttt{json} format but also to the naming conventions, style, and level of detail in function and parameter descriptions. Notably, based on our manual inspection, the style of the generated tool definitions is indistinguishable from that of the original function(s).
\vspace{-0.1in}
\paragraph{\textnormal{(3)}} 
In rare cases, a generated tool was found to be strictly functionally equivalent to the original one, despite differences in name, description, or parameter order (see Appendix~\ref{sec:appendix-c}). We eliminate such cases by (a) concatenating the original tool properties into a "signature," and (b) filtering out any newly generated tool whose "signature" exceeded a predefined similarity threshold to the original tool, as measured via cosine similarity of their embeddings, computed using the sentence-transformers module \citep{reimers-2019-sentence-bert}.

Table~\ref{tbl:toolkit-expansion-example} presents example original request (and its tool), along with the expansion process: additional (related but not strictly identical) request variants, and additional tools, fulfilling those additional requests. %Table~\ref{tbl:datasets_stats} further provides statistics on the expanded dataset. 
The mean number of tools in the expanded toolkit is 5.6 compared to the 2.7 (seemingly unrelated) tools in the original BFCL dataset, meaning that three semantically-related functions were added on average to each one of the 200 testcases. Next, we evaluate the FC performance of multiple agents using the generated benchmark.

\section{Agentic FC Robustness Evaluation}
\label{sec:experiments}

\begin{table*}[h!]
\centering
\resizebox{1.0\textwidth}{!}{
\begin{tabular}{l|ccc||cccc}
\multicolumn{1}{c|}{} & \multicolumn{3}{c||}{robustness evaluation} & \multicolumn{4}{c}{exp. toolkit + orig. query: error analysis (\%)} \\ \hline
model (agent) & original & \linebrcellc{orig. toolkit\\reph. query} & \linebrcellc{exp. toolkit\\orig. query} & \linebrcellc{wrong\\syntax} & \linebrcellc{wrong\\function} & \linebrcellc{wrong num\\of functions} & \linebrcellc{wrong param.\\assignment} \\ \hline
Llama3.1-70B            & 0.965 & 0.825 (-15\%) & 0.925 (-4\%) & 0.00 & 0.45 & 0.10 & 0.45 \\
Llama3.3-70B            & 0.945 & 0.785 (-17\%) & 0.905 (-4\%) & 0.00 & 0.23 & 0.46 & 0.31 \\
DeepSeek-V2.5           & 0.965 & 0.835 (-14\%) & 0.950 (-2\%) & 0.00 & 0.56 & 0.00 & 0.44 \\
Qwen2.5-72B             & 0.975 & 0.850 (-13\%) & \textbf{0.965 (-1\%)} & 0.00 & 0.29 & 0.00 & 0.71 \\
Granite3.1-8B-instruct  & 0.945 & 0.770 (-19\%) & 0.870 (-8\%) & 0.09 & 0.50 & 0.18 & 0.23 \\
\hdashline
Claude-3.5-Haiku        & 0.925 & 0.765 (-11\%) & 0.870 (-2\%) & 0.00 & 0.44 & 0.00 & 0.56 \\
Claude-3.5-Sonnet       & 0.915 & \textbf{0.845 (\hspace{0.15cm}-8\%)} & 0.890 (-3\%) & 0.00 & 0.29 & 0.00 & 0.71 \\
gpt4o-mini              & 0.925 & 0.765 (-17\%) & 0.870 (-6\%) & 0.26 & 0.42 & 0.00 & 0.32 \\
o1-mini                 & 0.905 & 0.770 (-15\%) & 0.885 (-2\%) & 0.33 & 0.27 & 0.00 & 0.43 \\
\end{tabular}
}
%\vspace{-0.05in}
\caption{Agentic FC robustness evaluation results. Models' AST performance drop is evident when rephrasing the original query, and also when using the original query with extended toolokit (left); relative percent drop is specified in brackets. Failures stemming from toolkit expansion vary mostly between wrong function selection and wrong parameter assignment (right). The best result in a column (the lowest performance drop) is boldfaced. %A similar distribution of failures stemming from user request rephrasing is reported in Appendix~\ref{sec:appendix-c}.
}
\label{tbl:results-summary}
\end{table*}

\subsection{Experimental Setup}

\paragraph{Models}
We evaluate several top-performing LLMs from the \href{https://gorilla.cs.berkeley.edu/leaderboard.html}{BFCL leaderboard}, both API-accessible and locally hosted, as FC agents. Closed models include GPT4o-mini and o1-mini,\footnote{\url{https://platform.openai.com/docs/models}} as well as Claude-3.5-Haiku and
Claude-3.5-Sonnet.\footnote{\url{https://www.anthropic.com/claude}} Locally hosted models include Llama3.1-70B and its more advanced version Llama3.3-70B \citep{dubey2024llama}, Granite3.1-8B-instruct \cite{granite2024granite}, DeepSeek-v2.5 \citep{deepseekv2}, and Qwen2.5-72B \citep{qwen2.5}.

\paragraph{Evaluation Approach}
BFCL employ a two-phase FC evaluation approach: (1) assessment of the generated tool call through the tree-matching abstract syntax tree (AST) methodology, and (2) evaluation of the tool execution in a simulated environment \cite{patil2023gorilla}. Our focus in this study is the evaluation of FC construction provided interventions in its input; we, therefore, adhere to the first evaluation phase -- namely, AST. A robust agent will generate correct function call regardless of the precise request wording and of its toolkit size: "thin" (as it comes with the original benchmark), or expanded, simulating a shortlister selection.

\subsection{Experimental Results}

We report AST averaged over the 200 dataset examples, including three variants: (a) the original version, (b) original ("thin") toolkit + rephrased user request, (c) expanded toolkit + original user request. Table~\ref{tbl:results-summary} (left) reports the results. Several insights can be drawn from the figures:

\paragraph{FC Evaluation Approach Weakness(es)}
A notable (and somewhat unexpected) drop occurs when evaluating the original toolkit on a rephrased request. Closer examination of errors reveals a significant weakness in the common approach to FC evaluation -- specifically, in handling arguments that can accept several equivalently valid values (e.g., named entities). Consider the request: "What is the humidity level in \texttt{Miami,Florida} in the upcoming 7 days?". The expected response includes the function \texttt{weather.humidity\_forecast()} and validates its \texttt{location} parameter by exact match to one of the predefined values: ["Miami", "Miami, Florida", "FL"]. When the request is rephrased as "How will the humidity levels change over the next seven days in \texttt{Miami,FL}?", agents assign the value "Miami, FL" to \texttt{location}, which does not match any of the (incompletely) listed options. 

Further systematic analysis of error types distribution reveals that 70--90\% of errors indeed stem from mis-match in parameter value assignment. We conclude that the majority of failures in this case can be attributed to the \textit{evaluation approach drawback} rather than agents' sensitivity.

We argue that this issue could potentially be mitigated by applying \textit{semantic similarity} instead of \textit{exact match}. Indeed, recent studies adopt a more holistic approach to evaluation of a constructed function call; e.g.,  \citet{zhong2025complexfuncbench} who use multi-dimensional matching strategy, including FCs' embeddings similarity and LLM-as-a-Judge matching, ensuring a generated tool call meets its semantic requirements. We leave the exploration of this mitigation strategy in the context of BFCL evaluation framework to future work. 

%Additional, albeit less frequent errors, reflect objective weaknesses of the models in constructing FCs that comply with tools requirements. As an example, rephrasing ...

\paragraph{Agents' Sensitivity to Toolkit Expansion}
Evidently, expanding an agent's toolkit with a set of related functions caused performance degradation across the board (Table~\ref{tbl:results-summary}, left). Here, objective agent failures span a range of error types: wrong function selected, wrong number of functions generated (typically two instead of one), wrong parameter assignment to a correctly-selected function, parameter hallucinations, etc. As an example, in response to the request "What is the ranking of Manchester United in Premier League?", an agent with the expanded toolkit produces \texttt{football\_league.ranking("premier league")}, retrieving the complete ranking table of the league, instead of the more appropriate \texttt{sports\_ranking("Manchester United", "premier league")}, answering the query. 

Table~\ref{tbl:results-summary} (right) presents error breakdown for agents in this study in the expanded toolkit scenario, showing the proportion of each error type within the set of failures stemming from toolkit expansion. While no clear pattern dominates, it is evident that agents struggle with both accurate function selection and parameter assignment.

Finally, expanding an agent's toolkit with additional functions occasionally caused models to "repair" some of their original (baseline) failures in a few cases. Interestingly, this observations highlights the stochastic, generative nature of LLM agents, where seemingly unrelated changes to a model context may entail different output.

\section{Conclusions and Future Work}
\label{sec:conclusions}

We focus on two aspect of robustness, capturing input variations that can be expected in real-world agentic deployments: (1) meaning-preserving rephrasings of user requests and (2) agent's toolkit expansion to include a set of semantically related tools that are likely to be shortlisted by a selection module. We build a benchmark dataset, evaluate the robustness of several SOTA LLM agents, and discuss the breakdown of failures. %We further discuss the breakdown of failures, highlighting (among others) prominent weaknesses of the existing agentic function calling evaluation benchmarks.

Our future work includes testing the robustness of agentic FC with additional and diverse datasets. Moreover, it has been shown that LLMs can be easily distracted by larger context \citep{shi2023large, levy2024same}. We plan to extend the set of experiments to scenarios were agent's toolkit is expanded also with non-relevant tools, to compare the performance against the current setting. 

\section{Limitations}
\label{sec:limitations}

While our study provides valuable insights into measuring agents' robustness in the function calling scenario, it has several limitations. First, we evaluate our approach on a single dataset, sufficient for the focused contribution of a short paper, but requiring extension to additional datasets for a broader analysis. Second, our toolkit expansion scenario relies on multiple LLMs to generate related requests and corresponding tools, a time-consuming process currently performed offline. We are actively exploring ways to streamline this pipeline for improved efficiency and usability.
\section{Ethical Considerations}

We use publicly available datasets to study the robustness of agentic function calling. We did not make use of AI-assisted technologies while writing this paper. We also did not hire human annotators at any stage of the research.

\section*{Acknowledgements}
We are deeply grateful to Michal Jacovi for her invaluable assistance in carrying out this study. We would like to thank Guy Uziel for his feedback on earlier versions of this paper. Finally, we are thankful to our anonymous reviewers for their useful comments and constructive feedback.

\bibliographystyle{acl_natbib}
\bibliography{anthology, custom}

%\clearpage
\section{Appendices}
\label{sec:appendix}

\subsection{Prompt for Request Rephrasing}
\label{sec:appendix-a}

We used the following prompt for generating \textit{strictly meaning-preserving} request rephrasing with the Llama3.1-70B model \citep{dubey2024llama}:

\vspace{0.08in}
\noindent 
\texttt{SYSTEM: You are a helpful assistant helping rephrasing user requests, while accurately preserving their meaning, including numbers and names if exist. Do not answer the requirement, just produce another one that is identical in meaning but is phrased differently. Produce ONLY the rephrased requirement, without further thoughts or explanations. Consider the example below:}

\vspace{0.08in}
\noindent \texttt{USER: Can I find the dimensions and properties of a triangle, if it is known that its three sides are 5 units, 4 units and 3 units long?}

\vspace{0.08in}
\noindent \texttt{ASSISTANT: What are the dimensions and properties of a triangle whose three sides are 5, 4 and 3 units long?}

\subsection{Prompt for Similar Requests Generation}
\label{sec:appendix-b}

We used the following prompt for generating \textit{closely related but different} request with the Llama3.1-70B model \citep{dubey2024llama}:

\vspace{0.08in}
\noindent \texttt{SYSTEM: You are a helpful assistant introduced with the following user query. 
Create a very similar query that refers to a very similar user need and is likely to be implemented in an enterprise as part of the same project. The new query should introduce one or two additional distinct parameter types. It should differ from the original query in a sense that a function that can be used to fulfill the original query is not fully appropriate for the new one and vise versa. As an example, generating 'Book a single room for two nights at the Hilton Hotel in Chicago' per the original query 'Book a double room for three nights at the Marriott hotel near OHare Airport in Chicago', is not sufficient since both queries can be answered using the same function call, invoked with different parameters. The query should contain all information needed for its computation. For instance, 'What is the capital of Brazil?' is a good query, while 'What is the capital of a country provided by user?' is not since one cannot generate a function call and populate its arguments using the info in the query alone. Output the newly generated query only, without explanation or interpretation. Consider the examples below:}

\vspace{0.08in}
\noindent \texttt{USER: I need the schedules of matches happening on February 28, 2024.}

\vspace{0.08in}
\noindent \texttt{ASSISTANT: I need the schedules of the college league matches happening during the winter 2024 season.}

\begin{comment}
\vspace{0.08in}
\noindent \texttt{USER: I need the schedules of matches happening on February 28, 2024.}

\vspace{0.08in}
\noindent \texttt{ASSISTANT: I need the schedules of the college league matches in Toronto happening on February 28, 2024.}
\end{comment}
\noindent \texttt{...}

\subsection{Example of Syntactically Different but Semantically Equivalent Tools}
\label{sec:appendix-c}

Although rare, distinct, yet functionally equivalent tools, pose a challenge for accurate evaluation, since the "labeled" BFCL data contains only one of these functions. As an example, the tool

\vspace{0.08in}
\noindent \texttt{sentence.translate(sentence: string,\\ \noindent from: string,\\ \noindent to: string)}

\vspace{0.08in}
\noindent is functionally equivalent to

\vspace{0.08in}
\noindent \texttt{translate\_sent(orig\_language: string,\\ \noindent target\_language: string,\\ \noindent sentence: string)}.

\vspace{0.08in}
As described in Section~\ref{sec:dataset}, we concatenate function name and description, as well parameter names and descriptions into a tool "signature", and filter out generated tools exhibiting cosine similarity higher than a predefined threshold to the original one, aiming at a toolkit with distinct functions. The similarity threshold was set to 0.8.

%A single test case  (out of 50) contained two equivalent tools per manual inspection of one of the authors, after the filtering procedure.

\begin{comment}
\subsection{Failure Distribution in the Case of Rephrased User Request}
\label{sec:appendix-c}

\begin{table}[h!]
\centering
\resizebox{1.0\columnwidth}{!}{
\begin{tabular}{l|cccc}
model (agent) & \linebrcellc{wrong\\syntax} & \linebrcellc{wrong\\function} & \linebrcellc{wrong num\\of functions} & \linebrcellc{wrong param.\\assignment} \\ \hline
%llama3.1-70B            & 0.00 & 0.45 & 0.10 & 0.45 \\
Llama3.3-70B            & 0.00 & 0.23 & 0.46 & 0.31 \\
DeepSeek-V2.5           & 0.00 & 0.56 & 0.00 & 0.44 \\
Qwen2.5-72B             & 0.00 & 0.29 & 0.00 & 0.71 \\
Granite3.1-8B-instruct  & 0.09 & 0.50 & 0.18 & 0.23 \\
\hdashline
%claude-3.5-haiku        & 0.00 & 0.44 & 0.00 & 0.56 \\
Claude-3.5-Sonnet       & 0.00 & 0.29 & 0.00 & 0.71 \\
gpt4o-mini              & 0.26 & 0.42 & 0.00 & 0.32 \\
o1-mini                 & 0.33 & 0.27 & 0.00 & 0.43 \\
\end{tabular}
}
%\vspace{-0.08in}
\caption{The distribution of failures stemming from user request rephrasing; the vast majority of errors fall into the wrong parameter assignment category -- the issue we attribute to the inherent weakness of the FC slot filling evaluation approach (i.e., exhaustively listing all possible parameter values).}
\label{tbl:results_p_req_robustness}
\end{table}

\end{comment}

\end{document}